\documentclass[sigconf]{acmart}

\AtBeginDocument{%
  \providecommand\BibTeX{{%
    \normalfont B\kern-0.5em{\scshape i\kern-0.25em b}\kern-0.8em\TeX}}}

\copyrightyear{2020} 
\acmYear{2020} 
\setcopyright{acmcopyright}\acmConference[PervasiveHealth '20]{14th EAI International Conference on Pervasive Computing Technologies for Healthcare}{May 18--20, 2020}{Atlanta, GA, USA}
\acmBooktitle{14th EAI International Conference on Pervasive Computing Technologies for Healthcare (PervasiveHealth '20), May 18--20, 2020, Atlanta, GA, USA}
\acmPrice{15.00}
\acmDOI{10.1145/3421937.3421971}
\acmISBN{978-1-4503-7532-0/20/05}

\usepackage{color,soul}
\usepackage[colorinlistoftodos]{todonotes}
\begin{document}

\title{A Framework for Addressing the Risks and Opportunities In AI-Supported Virtual Health Coaches}

\author{Sonia Baee}
\affiliation{%
  \city{University of Virginia}
}
\email{sb5ce@virginia.edu}

\author{Mark Rucker}
\affiliation{%
  \city{University of Virginia}
  }
\email{mr2an@virginia.edu}

\author{Anna Baglione}
\affiliation{%
  \city{University of Virginia}
}
\email{ab5bt@virginia.edu}

\author{Mawulolo K. Ameko}
\affiliation{%
  \city{University of Virginia}
 }
\email{mka9db@virginia.edu}

\author{Laura Barnes}
\affiliation{%
  \city{University of Virginia}
  }
\email{lb3dp@virginia.edu}

\renewcommand{\shortauthors}{Baee and Rucker, et al.}

\begin{abstract}
Virtual coaching has rapidly evolved into a foundational component of modern clinical practice. At a time when healthcare professionals are in short supply and the demand for low-cost treatments is ever-increasing, virtual health coaches~(VHCs) offer intervention-on-demand for those limited by finances or geographic access to care. More recently, AI-powered virtual coaches have become a viable complement to human coaches. However, the push for AI-powered coaching systems raises several important issues for researchers, designers, clinicians, and patients. In this paper, we present a novel framework to guide the design and development of virtual coaching systems. This framework augments a traditional data science pipeline with four key guiding goals: \textit{reliability}, \textit{fairness}, \textit{engagement}, and \textit{ethics}. 
\end{abstract}



\maketitle

\section{Introduction}

Healthcare costs are increasing dramatically due to disproportionate consumption of healthcare resources by patients. In the past decade, the public has become more health-conscious. This has increased the demand for consumer health information technology which allows individuals to monitor and manage their health and well-being and potentially alter their health-related behaviours. Health coaching systems can contribute to improved patient self-management while reducing costs due to increased scalability and availability of the use of human health coaches. In particular, AI-driven \textit{Virtual Health Coaches (VHCs)} can process information in real time and provide continuous health advice, opening new treatment avenues. With the evolution of VHCs, however, new problems have also arisen regarding safe handling of user data and ethical responses to challenging questions from patients, among other topics. In this paper, we propose a novel framework for guiding the design and development of AI-driven VHCs.

\section{Framework}
\begin{figure}
    \centering
    \includegraphics[width=0.3\textwidth]{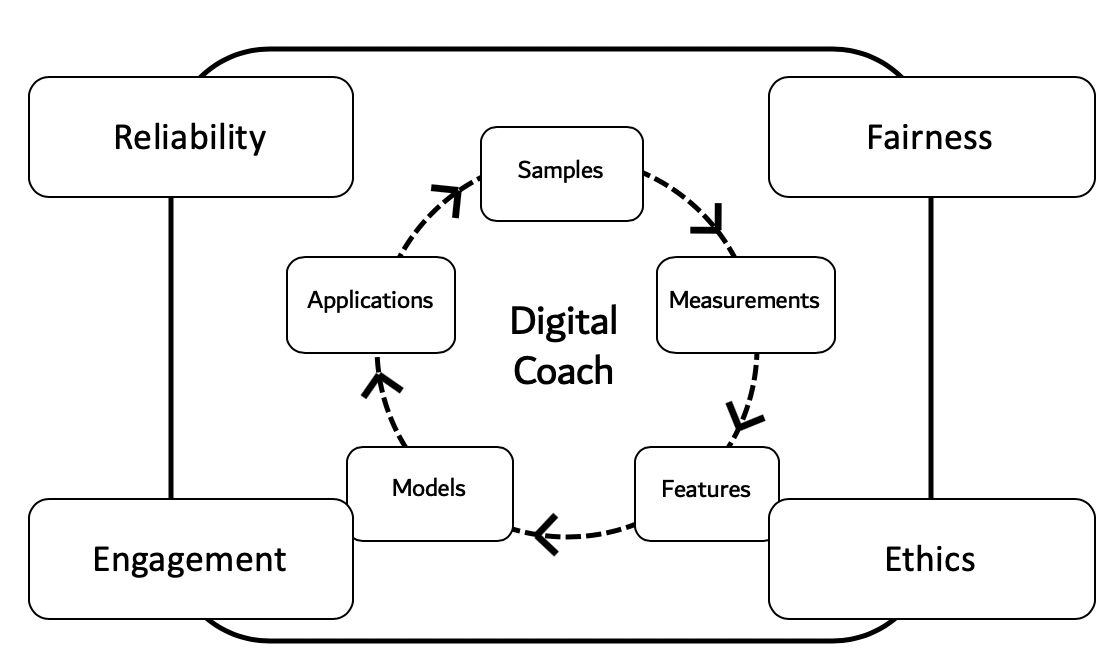}
      \caption{The figure shows four main domains of a successful virtual health coach throughout a data science pipeline.}
      \label{fig:framework}
\end{figure}

Our framework, shown in Figure~\ref{fig:framework}, proposes that the traditional data science pipeline (e.g. sample collection and measurement, feature extraction, modeling, and application) be reworked for VHCs. Namely, a successful VHC must prioritize four main goals at each stage of this pipeline: \textit{Reliability}, \textit{Fairness}, \textit{Engagement}, and \textit{Ethics}. In this section, we explore each of these goals in relation to VHCs.

{}

\subsection{Reliability}

\textit{Reliability} ensures that AI-powered VHC systems behave as intended. It is essential to consider and address the safety of a VHC as a safety-critical application. Reliability is often dependent on how data, or samples, are sourced. Data is hard to collect in healthcare applications and, when obtained, is often fraught with issues such as missing data and sample bias. These issues can lead to possible harmful behavior caused by bad extrapolations from limited samples~\cite{amodei2016concrete}. One such example might be a virtual dietary coach that recommends eating broccoli and cheese at night because the training sample was biased towards a subgroup of people who reported this food to be most effective for weight loss, thus ignoring differentiating factors such as wealth and culture. To build a reliable VHC will require a serious consideration of potential missing sample data, which could compromise the utility of the system to the user.

Measurement errors can occur even in carefully controlled sampling environments~(e.g., randomized control trials). Such errors in health coaching systems might result from sensor miscalibration~(e.g. accelerometer, for applications designed to monitor a patient's affect passively) or self-report error. Research in passive recognition of affect~\cite{ameko2018cluster} has used self-reported data from an ecological momentary assessment~(EMA) study to derive the most predictive features associated with affect. However, self-reported data, such as affect reported on a scale~(e.g, 0-10) is error-prone. Some participants might under-report their affect while others will over-report it. The true underlying affect is often hard to measure resulting in measurement errors, which in turn undermine inferences made from the data~\cite{adams2019learning}.

AI models for VHCs are often trained on samples which are not representative of the target population. Features that are predictive in the training set might not be relevant in the testing environment because of distributional shifts in domains. There is a growing literature base in learning domain-invariant feature representations that allow AI models to be reliable in new environments~\cite{hsu2018extracting}. Employing techniques developed in these models for VHCs will enhance reliability. Also because most machine learning models assume that data is independent and identically distributed~(IID), they are not directly adaptable to real-world phenomena since the data distribution varies from context to context over time. Models need to be robust to changes in distributions~\cite{subbaswamy2018counterfactual} and adaptive to individuals. People are different and largely moulded by their cultures; this creates a natural heterogeneity in the user population. Furthermore, people also might change over time as a function of their environments or co-morbidities. AI models must be adaptive to new domains, e.g. from person to person or cultural group to another. 

Modeling assumptions also need to be robust to small perturbations in the environments. They must be able to detect distributional changes in a patient's personal model and quickly adapt to these changes (e.g. detect if a person is experiencing additional stress by analyzing vocal tone changes in real time, if possible, and attempting to calm the patient). Generative adversarial networks~\cite{goodfellow2014generative} have received a lot of attention in recent years because they provide a framework for training machine learning models that are robust against adversarial attacks and noise, and are thus reliable in the deployed environment. Such properties should be required for AI models used in next generation VHC systems, given the current sensitivity of sensing devices to manipulation by bad actors. 
    
It is of critical importance that VHCs have safety mechanisms built in so that they do not abuse humans in application. Abuse of users is a learned behavior that can result if bad actors contribute abusive messages to the training sample pool. There have been a few instances over the last years where some agents went rogue after being subverted by online trolls and began tweeting racist propaganda~\cite{wolf2017we}. Requests from users to end communication should have a built in protocol to end the chat, preventing the agent from harassing or spamming a user. Further, language filters which detect potentially abusive language toward a user should be standard features of any VHC. Ultimately, from a users perspective, reliability must uphold cooperative behavior for a VHC to be useful.





%


\vspace{-1mm}
\subsection{Fairness}
In the future, VHCs built on machine learning~(ML) models must seriously consider issues of \textit{fairness}. We argue that a fair VHC is one which considers both social context and overall outcomes in all five phases of our framework to avoid \textit{disparate outcomes}. A similar view is held by some in the field of fair machine learning \cite{selbst2019fairness} and is in contrast to the view that fairness can be attained by simply optimizing for appropriate model metrics \cite{yao2017beyond}. 

Data samples are core to training ML models; if samples do not accurately represent a target population then there is a higher risk for disparate outcomes. Furthermore, if a VHC is designed to continue learning while online, then sampling becomes even more important since a coach which is initially trained on appropriately sampled data may, over time, learn from improperly sampled data. One well-known example of this risk is the chatbot Tay\cite{wolf2017we}. Due to malicious human interference, Tay's training data was biased towards racist and sexist conversations after Tay had been put online.

Inaccurate sensor measurements~(either in the training phase or deployment phase) also pose the risk of disparate outcomes since sensor readings can directly influence a VHC's behavior. This is true whether the sensor under consideration is a passive sensor (e.g., a smartphone accelerometer) or an active sensor (e.g., self-reported data). For instance, women have been shown to report pain more frequently than men~\cite{bartley2013sex}. Based on this, it is easy to imagine how a VHC may over-treat women's pain or under-treat men's pain without careful considerations. To the authors, these kinds of problems are matters of fairness due to the belief that they will have a tendency to impact specific subgroups of the population.

The choice to include or exclude any feature in a model carries implicit information about what is important for making decisions. For example, the criminal justice corpus contains considerable debate over which features are important when predicting recidivism risk \cite{dressel2018accuracy} (e.g., is it important to consider race when predicting recidivism risk?). Unfortunately, achieving feature fairness is rarely as simple as including or excluding a feature since secondary features may have a high correlation with problematic constructs. That is, one may choose to exclude race from a recidivism model but include zip code which can be highly correlated in segregated communities.

In general, model selection can impact fairness via three mediators: (1) \textit{features}, (2) \textit{freedom}, and (3) \textit{objectives}. Some models implicitly consider all feature interactions, potentially introducing unfair criteria. Models with a high degree of freedom will be more sensitive to problems in the training data. Finally, every model will have its own set of well-known objective functions; the selected model should have an objective that aligns well with the targeted application.

Systems designers must carefully consider the metrics, subgroups and contexts that will be monitored to evaluate effectiveness. For example, \cite{obermeyer2019dissecting} has shown that using cost as an outcome metric in hospitals can lead to disparate health outcomes for white and black patients. Further, \cite{chouldechova2017fair} has shown that parole hearings that use recidivism rates to evaluate outcomes rather than re-offense rates render any argument about fairness suspect. Ultimately, the application phase is the only phase that will determine if a virtual health coach is fair, as all issues of fairness in the previous four phases are considered simply to generate a fair application.

\vspace{-1mm}
\subsection{Engagement}

An \textit{engaging} VHC is one which holds' the user's attention and provides value to the user in both the short and long term. To this end, VHCs should rely on samples with similar challenges and goals to those of the target population. For instance, a virtual coach designed to improve mental health outcomes in anxious populations is more likely to be engaging if it is trained on a dataset of text messages and heart rate variations in a sample of anxious individuals. 

Engagement tends to be a minimal concern in the measurement phase of data collection, especially in contexts where the majority of data is gathered using unobtrusive passive sensing (e.g., automatic capture of GPS location). One notable exception, however, is experience sampling. Designers must be careful not to overburden a VHC user with an abundance of notifications in the quest to gather biometric or self-report data, as too-frequent or ill-timed notifications may disrupt the user's routine and increase the risk of abandonment. For instance, a user who works the night shift would be frustrated to receive a request to answer an EMA at 8am after just arriving home and lying down to sleep. Determining a reasonable maximum threshold for the number of data requests sent per day and optimizing the delivery of these requests based on the user's schedule remain open research questions. Such study scheduling procedures should be developed in consultation with real users. 

Modeling engagement has historically relied on metrics such as  \textit{dwell time} (i.e. time spent on a particular task), click count (i.e. the number of times a user clicks on a particular element) ~\cite{wu_returning_2017, zou_reinforcement_2019}, and frequency of application usage~\cite{cheung_engagement_intellicare_2018}. What features appropriately represent engagement will differ from VHCs in different domains~(e.g. mental health, dietary). Regardless of which features a VHC relies on, researchers have cautioned against a ``quantity over quality'' approach when examining population-level engagement. For example, counting the number of apps installs at the population level serves as a poor measure of engagement, as an app may be installed yet remain unused or used sparingly ~\cite{bohmer_appfunnel_2013}.

In practice, mHealth interventions are plagued by high attrition~(dropout) rates~\cite{torous_dropout_2020}. An engaging VHC must be prepared to leverage strategies that sustain engagement both over the short and long term. Various strategies for increasing engagement and retention during the application phase have been proposed, including the use of incentives~\cite{rabbi2018toward} or \textit{gamification}. Gamification is defined as ``the use of game design elements in non-
game contexts''~\cite{deterding_gamification_2011} and has seen recent adoption in a variety of eHealth interventions ~\cite{brown_eHealth_gamification_2016, pramana_anx_gamification_2018}. In the context of virtual coaching, a user who adheres to her treatment protocol for anxiety by checking in with her VHC at least once per week could, for instance, be awarded a badge for each week completed. 

In addition to cultivating and sustaining engagement, a robust VHC must be able to respond appropriately to a user, and thus must maintain at least a crude representation of user context. A successful VHC might, for instance, maintain a live engagement score during user conversations and change its approach when the score dips below a certain threshold. For example, a user who starts giving one-word answers to a difficult question (say, about an adverse childhood experience) would trigger a dipping score alert, and the VHC could respond by either posing a different question to the user or by encouraging the user not to be afraid to explore the existing question, much like a real therapist might do. While this process for determining engagement is indeed binary and quite mechanical, it is necessary until AI-powered VHCs gain a sophisticated understanding of user context.








\vspace{-1mm}
\subsection{Ethics}
\textit{Ethics} considers the potential impact of a technology on society. Examining ethical concerns can help researchers identify, track, and mitigate both intended and unintended consequences of VHCs~\cite{metcalf2019owning}. The topic of ethics in a VHC is complex and spans a wide area including privacy, data ownership, and transparency.

Training and intervention population sample selection can perpetuate bias or pose undue risks, in some cases. For example, a system trained on data from only men or only white individuals could perpetuate racial or gender disparities when deployed in the real world.
Bias and ineffective solutions of algorithmic fairness threaten the ethical obligation to avoid or minimise harms to users of VHC.
Or, perhaps the research phase is invasive or risky and so only financially vulnerable populations volunteer. Finally, it should be considered how the data collected will be protected and how it could harm the training and intervention sample populations were it to be compromised. Historical documents such as the Nuremberg Code have been developed to address these issues and should guide the design process.

Data measurement methods also pose key ethical challenges. For example, measuring blood glucose might require users to ``harm'' themselves by pricking a finger for blood or sleep sensors that may violate the privacy of overnight guests if placed on a bed. Using expensive and proprietary sensors could exclude certain target populations. Each of the above examples is not enough to end a study, but for researchers are realities that can easily be overlooked.

Feature selection for VHCs requires both design-time and intervention~-time decisions. Researchers need to proactively consider the privacy risks associated with collecting certain features. If possible, features that minimize this risk should be preferred. For example, researchers might choose to track the time spent in bed rather than the amount of movement in bed which could be invasive. On the reactive side, researchers should have a plan for how to respond when at-risk features are observed (e.g., poor vitals, a sudden spike in a diagnosed mood disorder or failure to adhere to prescription regimens).

Modeling choices are often the source of real-world errors. Unfortunately, because many such systems are ``black boxes", the reasons for these errors are often not easily accessed or understood by humans. Therefore, models that emphasize easily-interpretable features~(for both researcher and user) should be prioritized. Additionally, models must be constructed with safeguards in place to ensure that inappropriate recommendations are never made~(e.g. further restricting one's eating while battling anorexia, staying with an abuser, or entertaining suicidal thoughts) and to connect a user with the appropriate resources if the VHC cannot provide the necessary assistance.

Researchers must embed ethics into the formulation of their research from day one, and must understand the potential harms of algorithmic injustice in society. Issues such as algorithms reproducing existing patterns of discrimination in society and disproportionately harming already vulnerable populations, are critical considerations for the application phase of digital coach development~\cite{bostrom2014ethics}. Furthermore, researchers or businesses who build VHCs should provide clarity about who owns what data and should clearly explain their terms of service. This, of course, is another battle for ethical designers, given the lack of transparency in terms of service agreements~\cite{kluge2020artificial}. Ethical considerations have rarely been so integral and essential to maximising success of a technology both empirically and clinically.

\section{Conclusion}

We have proposed a novel framework for the design and development of AI-driven VHCs, and have argued that \textit{reliability}, \textit{fairness}, \textit{engagement}, and \textit{ethics} are critical considerations for the VHC of the future. A VHC designed with these tenets in mind will help build trust between VHCs and users. When building a successful human-centered technology, the most important step is aligning the product with user needs. Therefore, using a human-centered design approach when developing a VHC is decisive for assessing the true impact of its usages, predictions, recommendations, and decisions. AI-powered VHCs are largely a reflection of the patterns inherent in their training data. Describing the scope and coverage of these patterns to stakeholders like clinicians and patients in a clear, concise manner is an essential piece of the VHC development process, as it establishes transparency and builds trust. Further, designers and researchers should also communicate limitations to users when possible. Users should know what they are consenting to when using VHCs. In particular, they should know that VHCs cannot fully replace a human coach at this time. 

We acknowledge that this area is dynamic and evolving, and we will approach our work with humility, a commitment to internal and external engagement, and a willingness to adapt our approach as we learn over time. Our framework is not comprehensive by any means. VHCs are still in their relative infancy, and new issues that remain unaddressed by our framework are bound to arise. Researchers and clinicians should continue to learn from emerging exemplars of both successful and troubled VHCs to establish shared guiding principles for VHC development.

\bibliographystyle{ACM-Reference-Format}
\bibliography{reference}

\end{document}